\documentclass[letter,10pt,conference]{ieeeconf} 

\IEEEoverridecommandlockouts 
\usepackage{graphics} 
\usepackage{epsfig} 
\usepackage{mathptmx} 
\usepackage{times}
\usepackage{amsmath}
\usepackage{amssymb} 
\usepackage{comment}
\usepackage{xcolor}
\usepackage{makecell}

\usepackage{tikz}
\usepackage{color}
\usepackage[export]{adjustbox} 

\DeclareMathOperator*{\argmax}{argmax}

\title{\LARGE \bf
Quality-guided UAV Surface Exploration for 3D Reconstruction
}

\author{Benjamin Sportich¹, Kenza Boubakri¹, Olivier Simonin¹, Alessandro Renzaglia¹
\thanks{$^{1}$Inria, INSA Lyon, CITI, 69621 Villeurbanne, France. Email: {\tt\small fistname.lastname@inria.fr}}
\thanks{This work has been funded by the Agence Nationale de la Recherche (ANR), AVENUE project, grant ANR-22-CE33-0004.}
}

\begin{document}

\maketitle
\thispagestyle{empty}
\pagestyle{empty}

\begin{abstract}
Reasons for mapping an unknown environment with autonomous robots are wide-ranging, but in practice, they are often overlooked when developing planning strategies. Rapid information gathering and comprehensive structural assessment of buildings have different requirements and therefore necessitate distinct methodologies. In this paper, we propose a novel modular Next-Best-View (NBV) planning framework for aerial robots that explicitly uses an explicit reconstruction quality objective to guide the exploration planning.  In particular, our approach introduces new and efficient methods for view generation and selection of viewpoint candidates that are adaptive to the user-defined confidence objectives, exploiting the uncertainty encoded in a Truncated Signed Distance field (TSDF) representation of the environment. This results in informed and efficient exploration decisions tailored towards the predetermined objective. Finally, we validate our method via extensive simulations in realistic environments.  We demonstrate that it successfully adjusts its behavior to the user goal while consistently outperforming conventional NBV strategies in terms of coverage, quality of the final 3D map and path efficiency.
\end{abstract}

\section{Introduction}

Autonomous exploration for 3D reconstruction is a fundamental task in robotics, with critical real-world applications such as infrastructure inspection, mapping, environmental monitoring, and search-and-rescue missions \cite{quattrini2020}. In these scenarios, Unmanned Aerial Vehicles (UAVs) equipped with onboard visual and range sensors have proven essential for efficiently navigating complex environments and capturing aerial perspectives that enable comprehensive 3D reconstructions or a swift survey of an area.
Solving this problem usually involves selecting efficient sequences of viewpoints that maximize the expected information gain from new observations, which is particularly challenging in environments with complex structures and in the absence of any prior knowledge of their geometry. When using small UAVs, this task becomes even more difficult due to the limited computational resources typically available on these platforms, which require lightweight and efficient algorithms capable of operating in real time even on large-scale maps.

Despite the variety of applications, exploration and reconstruction strategies have often been considered as solving the same problem.  
However, while several applications require a balance between the two tasks, their objectives are often in conflict. The fast coverage needed for search and rescue applications is typically incompatible with the precise sensor measurements necessary for high-accuracy reconstruction. Similarly, the measured pace demanded by infrastructure inspection is not suitable for emergency use cases. 
For these reasons, we present a novel Next-Best-View (NBV) planning solution that explicitly takes into account the user's goal by defining a quality objective for the 3D reconstruction of the initially unknown and arbitrary environment to explore.
Depending on the initial fixed purpose, our method can prioritize speed over accuracy, accuracy over speed, or a balance of the two at every step of the strategy by leveraging a purpose-designed function.

The main contributions of this paper are as follows:
\begin{itemize}
    \item A new representation-based formalism reflecting reconstruction quality and completeness of the built map;
    \item A novel Next-Best-View approach relying on this formalism and able to adapt to the targeted quality both in the views generation and selection stage;
    \item An extensive evaluation in both indoor and outdoor realistic scenarios that demonstrates the competitiveness of our solution in terms of coverage, speed, and map accuracy.
\end{itemize}

\begin{figure}[t]
\centering
\includegraphics[width=0.78\columnwidth]{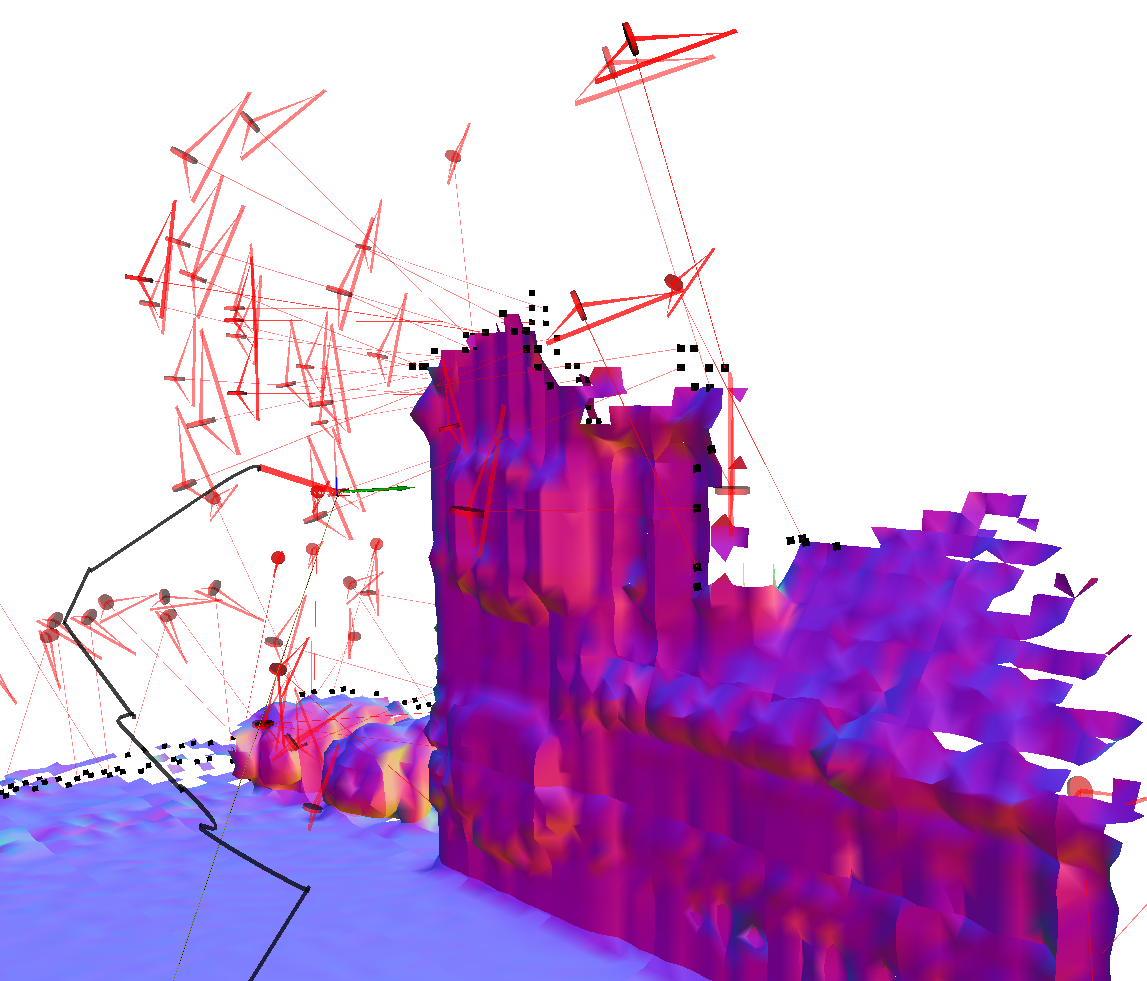}
    \caption{Illustration of an autonomous exploration and 3D reconstruction task with an aerial robot. Red points in the free space represent candidate viewpoints generated to select the next robot's position, black ones are surface frontiers within the sensor's range.}
\label{fig:simulation_overview}
\end{figure}

\section{Related Work}
\label{sec:rel_work}
Autonomously exploring an environment and obtaining an accurate 3D reconstruction of an unknown space are two very similar yet distinct problems that have been widely studied in recent years. In both cases, a standard way to tackle these problems is to adopt an iterative structure called Next-Best-View (NBV) planning: based on the available partial knowledge of the environment, the most promising future positions or portions of paths are selected using a set of criteria and proxies. Exploration-based solutions are typically evaluated on speed or volume coverage \cite{zhou2021}, while reconstruction-based approaches are evaluated mostly on the deviation from a ground-truth 3D mesh \cite{hardouin2023}. 
In both cases, most existing solutions, as  \cite{zhou2021}, \cite{hardouin2023}, \cite{Schmid2020}, leverage the seminal concept of frontiers, first introduced in \cite{yamauchi} for 2D environments, to guide the mission.
This concept was then extended to 3D and specifically for surface reconstruction, by introducing surface frontiers, which represent the border between empty and unknown space around the observed geometry. 
Other solutions are directly based on sampling-based approaches in the 3D space. Within this line, the solution presented in \cite{bircher2018} randomly samples the freespace using RRT and iteratively selects the most promising branch to explore.
Although fast and probabilistically complete due to random sampling, the quality of views and reconstruction performance of this type of method are typically lower than those of solutions that generate views directly from the surface geometry rather than the free volume. 
\cite{selin2019} and \cite{Schmid2020} combine both methods, using RRT to explore the freespace but sampling around the surface frontiers to exploit the environment structure. A different sampling-based solution that approximates a reward function based on past observations is then presented in \cite{renzaglia2019}.
Several solutions also consider predefined structures, such as a demi-sphere, to sample views around the target, whether an object \cite{delmerico2018} \cite{flow_nbv} or an outdoor environment \cite{peng2019}, assuming a domain with specific geometric properties. In our work, we propose a solution that generates views directly based on surface frontiers and the perceived scene geometry. Additionally, our experimental evaluation takes place in settings with complex topologies, without any hypothesis on their structure.

\subsection{Entropy \& Heuristics}
\label{sec:heuristics}
To select the Next-Best-View among a pool of view candidates, many evaluation functions have been proposed.
A common and reliable option is to leverage the concept of entropy by examining all the voxels in the sensor's field of view and assessing their individual information gain based on their occupation probability. In \cite{delmerico2018}, the authors propose and analyze eight different evaluation functions that combine distinct criteria and heuristic-based terms. Although performant, these methods are particularly expensive and often incompatible with the computational power embedded on a UAV platform. Thus, they are commonly restricted to object reconstruction with robot arms.
These limitations have led to the development of different NBV evaluation strategies for 3D exploration based on heuristics such as occlusion checks and visible frontiers counts in the field of views \cite{hardouin2023}.

To increase efficiency and reduce unnecessary travels, some approaches, such as  \cite{zhou2021}, \cite{hardouin2023}, \cite{tang2023}, try to compute an optimal order of visits of available views or sectors of the map (often relying on an asymmetric TSP model). This solution involves, however, heavy computational costs to obtain a still sub-optimal solution, since it remains based on partial information. A common and lighter alternative to optimize the robot's navigation relies on the use of heuristics that incorporate distance and angular costs, as proposed in \cite{yu2023}. This approach provides competitive results for a fraction of the cost of the TSP model.

\subsection{3D Representations}
The design of solutions is also strongly influenced by the adopted 3D representation, especially for high-quality reconstruction. Explicit representations, such as the octree-based occupation grid Octomap \cite{hornung2013}, can be used for surface reconstruction and navigation without post-processing. However, their computational cost quickly becomes very high in large environments. They have also been successfully used in NBV approaches designed for single-object reconstruction, as in \cite{delmerico2018}, \cite{flow_nbv}, with a more tractable cost, but showing some limitations in terms of accuracy of the final reconstruction. 
Lately, the use of Signed Distance Fields, implicit representations of the environment, has been largely preferred. More economical, they allow a greater fidelity through a finer discretization \cite{tsdf_foundation}. Truncated Signed Distance Fields (TSDF) use projective distances to obstacles within a sparse voxel grid and model uncertainty. On the other hand, Euclidean Signed Distance Fields (ESDF) use non-projective distance to obstacles and are particularly efficient for path planning \cite{chomp}. \cite{voxblox} and \cite{pan2022_voxfield}, building an ESDF and TSDF concomitantly, and \cite{fiesta} only creating an ESDF, are standard libraries for sensor input integration in the literature.
Notably, the solution presented in \cite{border2022} does not rely on any 3D representation and reasons directly on point clouds, but its definition does not allow perceiving empty space and requires specific processing for occlusion management. 
In this paper, we propose a representation-based framework exploiting the characteristics of SDF, as motivated in the next section. 

To our knowledge, our solution is the first to propose an NBV approach that adapts both view generation and selection to a required confidence level on the final map.

\section{Problem Formulation}
\label{sec:problem}
We consider an aerial robot equipped with a depth sensor with a limited Field-of-View (FoV) characterized by a maximum and minimum range $r_{max}$ and $r_{min}$, respectively. We assume access to a perfect localization system at all times in a reference frame that we will call the world frame. The robot's goal is to reconstruct the 3D surface of an unknown environment in the minimum time possible while taking into account predefined quality objectives. The environment is initially unknown and we represent the scene as a triangular mesh \( \mathcal{M} = \{T, O\} \) built on a pointcloud 
with mesh vertices \( O = \{o_1, o_2, \ldots, o_{|O|} \} \) and a set of faces \( T = \{t_1, t_2, \ldots, t_{|T|}\} \), where \( t_i \in V \times V \times V \), following the definition in \cite{peng2019}.
We call $Z$ the quality measure of the 3D reconstruction, based on the built map $M$, which will be later formally defined.
   Let $q \in SE(3)$ be the pose of the robot and $q_{0}$ its initial configuration. $q$ is defined by a position $s_q \, \in \, \mathbb{R}^{3}$ and orientation $\phi_q \, \in \, \mathbb{R}^{3}$. 

\textbf{Reconstruction problem:} Considering our UAV with initial configuration $q_0 \in Q$, the reconstruction problem asks to identify $S = (  \hat{Q}, \hat{P}_{\hat{Q}} )$, a set of poses and the corresponding collision-free paths which allow the agent to reach a predefined quality of reconstruction $Z^{*}$ in a minimum time.

The quality measure of the reconstruction and the corresponding minimum threshold can be tailored depending on the exact application and constraints of the problem. In this section, we will detail the design of our proposed $Z$ function. 

We consider a volumetric map $M$ of the environment based on a discretized voxel grid of the 3D space. A voxel, of size $d_s$, can refer to \textit{free}, \textit{occupied}, or \textit{unknown} space. We follow the general definition of surface frontiers given in \cite{hardouin2023}: 

\textbf{Definition 1 (Incomplete Surface Element):} 
An
\textit{Incomplete Surface Element} (\textit{ISE}), is a voxel $i \in M$ lying on the surface at a frontier, near both the unknown and empty space. Let $C$ be the set of all ISEs, a voxel $i \in C$ if and only if:
\begin{itemize}
    \item $i$ is \textit{empty}, 
    \item $\exists \, i' \in \mathcal{N}^6_i \; \text{s.t. $i'$ is \textit{unknown},}$
    \item $\exists \; i'' \in \mathcal{N}^{18}_i \; \text{s.t. $i''$ is \textit{occupied},}$
\end{itemize}
where $\mathcal{N}^6_i$ and $\mathcal{N}^{18}_i$ denote the 6- and 18-connected voxel neighborhoods of $i$, respectively. The definition of an empty, unknown, occupied voxel is dependent on the map representation and we will detail this further below.

\textbf{Definition 2 (Remaining Incomplete Surface):} Let $Q$ be the set of all collision-free configurations of an agent, and let $Q_c \subseteq Q$ be the set of all configurations from which an ISE $i \in C$ can be scanned. The remaining incomplete surface is then defined as:
\begin{equation}
C_{\text{rem}} = \bigcup_{i \in C} \{ i \mid Q_c = \emptyset \}.
\end{equation}

\subsection{Reconstruction Quality in TSDF } 
To represent the map, we use a Truncated Signed Distance Field (TSDF), where each voxel $i \in M$ is represented by an aggregated projected distance $d_i$ to the closest obstacle and a weight $w_i$, which serves as a measure of confidence intance.
Within a TSDF representation, a voxel $i$ is considered:
\begin{itemize}
    \item empty, if $w_i > 0$ and $d_i > d_s$ (voxel size),
    \item occupied, if  $w_i > 0$ and $d_i < d_s$,
    \item unknown, if $w_i = 0$.
\end{itemize} 

The choice of a TSDF representation is justified by several reasons. First, a TSDF implicitly represents the surfaces but explicitly models the confidence in the progressively built map, which is crucial in assessing the accuracy of the reconstruction online. In addition, the seminal work \cite{kinect} has extensively studied the impact of the sensors' uncertainty in reconstruction problems. These results are now commonly used to integrate the sensor input, as done for instance in \cite{voxblox} and \cite{pan2022_voxfield}, two widely-used TSDF implementations for UAV exploration and reconstruction.
It is important to note that the TSDF representation has been proved to be equivalent to the least squares minimization of squared distances between points on the range surfaces and points on the desired reconstruction under the specific conditions of an orthographic sensor and a range error independently distributed along the line of sight of the sensors  \cite{tsdf_foundation}. 
Hence, although the TSDF weights $w_i$ are not a direct geometric error metric, we propose to use them to define a function $Z(M)$ which acts as an efficient proxy for a quality measure of the reconstructed isosurface:
\begin{equation}
Z(M) = \xi (M) \, \bar{Z} = \xi (M) \sum_{i \in M_S}{ \frac{w_i}{ |M_S| }} 
\label{quality_function}
\end{equation}
where $M_S$ is the set of observed surface voxels in $M$ and $\xi (M) = 0 $ if $\exists \, i$ such that $i \in C$ and $i \notin C_{\text{rem}}$, and $ \xi (M) = 1 $ in all other cases. 
As a result, $ \xi (M) $ evaluates the completeness of the surface reconstruction and $\bar{Z}$ represents the average confidence in the obtained map. 
Contrary to a quality function incorporating directly distance errors or surface normals that can only be known a posteriori, our proxy function can be used online to evaluate a partial map. 

Using the function in equation~\eqref{quality_function} can however lead to overestimating the overall accuracy of a map, where some parts of unnecessarily high confidence compensate for lower-accuracy areas. We thus define the following modified quality evaluation function $Z_{sat}(M;w^{*})$:
\begin{equation}
Z_{sat}(M;w^{*}) = \xi (M) \, \bar{Z}_{sat} = \xi (M) \sum_{i \in M_S}{ \frac{\text{min(}w_i;w^{*} )}{ |M_S| }} 
\label{quality_function_2}
\end{equation}
where $w^{*}$ is a predefined confidence goal. 
Using this formulation, we can formulate our problem as finding the shortest path that maximizes the function (\ref{quality_function_2}) given a predefined confidence objective $w^{*}$. Note that $Z^*:=\text{max}(Z_{sat})=w^*$.

Although this optimization problem cannot be solved directly, the following section presents an online NBV planning approach that aims to achieve the same objective.

\section{Proposed Approach} \label{sec:approach}
To reconstruct or explore an unknown environment, Next-Best-Views solutions proceed iteratively: based on the partial map of the environment available at a given time, they select the viewpoint that will bring the most expected information and thus reduce mission time. Viewpoints are generally constructed around frontier points, borders between the unknown space and the known surface, that constitute intuitively highly informative viewing zones. Typically, the process consists of successively identifying frontier points, generating viewpoints around them, which exploits the structure of the known map, and finally selecting the next robot pose among all the potential destinations. 
Repeating this process involves moving from one locally optimal view candidate to another, according to considerations about the unknown surface and metric distance penalties to reduce back-and-forth motions. In this work, we introduce a novel way to determine locally optimal views by incorporating explicitly user-defined quality requirements in this procedure. To this end, we designed a new quality objective function to maximize, which is used at each step of the process to orient the exploration towards the chosen quality goal.

Our objective function $Z_{sat}(M;w^*)$ encompasses two criteria: complete exploration and high-confidence reconstruction. With a limited time, these objectives are at odds, as the former requires fast traversal of unknown spaces and the latter requires careful observation of sometimes complex geometry. Typically, these conflicting goals may also require different viewing distances of the target. A fast exploration strategy favors high-distance observations, maximizing unknown space sensing, while a reconstruction method requires closer observation points in search for accuracy. 
These complexities are reflected in our function design. By setting a quality objective $Z^{*}$ that informs both the generation and the evaluation of viewpoints, the user can calibrate the speed and reconstruction fidelity according to their preferences. A $Z^{*}$ tending to 0 would orient the planning towards an aggressive exploration strategy, while a value closer to 1 would make the exploration slower, careful and progressive.

\subsection{Use of $Z^*$ to Guide the Exploration}

Our objective function $Z_{sat}$, defined in eq.~\eqref{quality_function_2}, is built on the TSDF representation and exploits the associated confidence estimates encoded in the weights $w_i$. Leveraging the results presented in \cite{nguyen2012modeling}, the confidence of a given measurement is typically set to $1/d^2$, where $d$ is the measurement depth. For example, this is the case for the TSDF library Voxblox \cite{voxblox} that we adopt in our work. 
Based on this model, to obtain a confidence $w^*$ for a given surface voxel would be sufficient to observe it from a distance $d^*$ equal to:
\begin{equation}
    d^{*} = \min{ (\sqrt{ 1 / w^{*}}, r_{max} )} \,.
\end{equation}
On the map level, this relation implies that, if all surface voxels are observed at a distance of $d^{*}$ or less at least once, then $Z \geq Z^{*}$.
When $Z^{*}$ is close to 0, $d^{*}$ would correspond to the maximum range of the sensor, meaning voxels can be observed from any distance and still satisfy the quality requirements. On the other hand, if $Z^{*}$ is close to 1, then $d^{*}$ will be close to the sensor's minimum range. It is important to note that a given confidence level could also be achieved with multiple observations obtained from distances greater than $d^{*}$ (see \cite{voxblox} for more details on the weight update rule).
Although it is not possible to define a priori the optimal observation distance to reach a predefined $Z^*$ in the shortest possible time, we will use $d^*$ as our reference observation distance to guide both view generation and evaluation. In practice, we define a viewing distance interval $[d^*-\eta ; d^*+\eta]$ with $\eta>0$, and our strategy aims for a complete exploration via the generation and selection of views within this viewing distance interval.

\begin{figure}[t]
\centering
\includegraphics[width=.99\columnwidth]{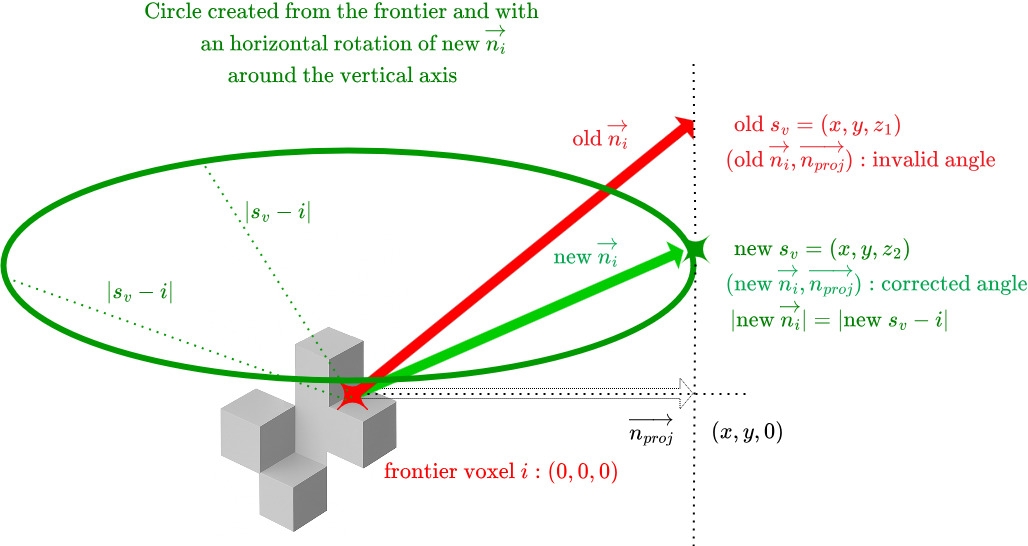}
    \caption{Vertical and horizontal rotation for the view generation. (
    The red and green arrows indicate, respectively, a direction from the frontier to a view candidate whose vertical angle is incompatible with the sensor's opening, and the corrected direction toward an adjusted view candidate. The green circle represents the sampling space of view candidates with a valid vertical angle and viewing distance to the frontier. \vspace{-0.2cm}}
\label{fig:view_generation}
\end{figure}

\subsection{View Generation}

When the robot starts its exploration or is about to reach the current goal position and orientation, the map is examined, and a view candidate is generated for every surface frontier (ISE voxel as in Definition 1) identified in the map. 
A view candidate $v \in SE(3) $ is defined by a position $s_v \, \in \, \mathbb{R}^{3}$ and an orientation $\phi_v \, \in \, \mathbb{R}^{3}$. 
In contrast to sampling-based approaches that sample randomly in freespace, frontier-based views exploit the geometry of the environment to propose efficient and safe observation poses. Three points are sampled along the normal of the surface frontiers points at different viewing distances $d$: in a close range $d <  d^*-\eta$, in an optimal range ($d \in [d^*-\eta ; d^*+\eta] $), and at long range ($d >  d^*+\eta$). Points within the optimal range, then close and long range, are checked for safety in this respective order. This order is in accordance with the satisfaction of the aimed quality threshold. Once a valid position is found, the process is stopped and moves to the next frontier.

In signed distance fields, the normal of a surface point is equal to the gradient direction at the position of the corresponding surface voxel. Typical gradient calculation methods of forward and central difference, although computationally efficient, do not perform well in an exploration context, because of the perturbation of unknown voxels surrounding the surface. We chose to implement the gradient calculation developed in \cite{hardouin2023}, which exploits all neighbors and incorporates uncertainty. The surface normal direction $n_i$ of the surface frontier $i$ is computed as:
\begin{equation}
    n_i = \sum_{j \in \mathcal{N}^{26}_i} w(j) \frac{ j - i}{|| j - i ||}
\end{equation}
with $\mathcal{N}^{26}_i $ the set of 26-neighbors of $i$ and $w(j) = -w_j$ if $j$ is occupied and equals to $w_j$ otherwise. 

If no admissible view is found due to the close vicinity of obstacles within the robot's radius or if the incidence angle is not aligned with the sensor FoV, the sampling direction $n_i$ is then rotated horizontally and/or vertically (see Fig. \ref{fig:view_generation}). Notably, if the vertical angle falls outside the FoV, the surface normal vector is rotated by the necessary angle in the corresponding plane to bring it back in the center of the field of view before performing a collision check. In this way, all the generated views are compatible with the robot's sensing capabilities. If for the frontier $i$, $s_v$ fails the collision check, we horizontally rotate the viewing direction $n_i$ around the vertical axis centered on the frontier $i$. Any new view candidate for $i$ of position $s_v'$ is sampled from the circle obtained through this rotation. This process ensures that neither the vertical angle nor the distance to the frontier ($||s_v'- i||=||s_v- i||$) are modified.
The set of generated candidate viewpoints is called $V$. 

\subsection{View Evaluation}
The view candidates generated in $V$ are then evaluated according to an information gain component $J_I$ and a navigation component $J_N$. The selected view $v^*$ is the view that maximizes the function $J$ that combines these two terms, i.e.: 
\begin{equation}
    v^* = \argmax_{v \in V} \; J(v) = \argmax_{v \in V} \;\alpha \, J_I(v) + \beta \, J_N(v)
\end{equation}
with $\alpha, \beta \in [0;1]$.
Considering the significant number of view candidates, an evaluation function
needs to both discriminate efficiently and be computationally tractable, as the size of $V$ can grow exponentially with the size of the map. 

\subsubsection{Information Gain Component}

Information gain evaluation functions can be sorted into two main categories: entropy-based and frontier-based. In Section \ref{sec:heuristics}, we discussed the prohibitive cost of entropy-dependent solutions for large environments in comparison to frontier-based methods. For this reason, our information gain model is based on frontiers, but does not limit itself to counting visible frontiers in the field of view. Our function is guided by the objective quality $Z^*$, which considers viewing distance and visibility probabilities due to occlusions to differentiate views.

A voxel $i$ is generally called \textit{occluded} from a view candidate $v$ if along the ray from $v$ to $i$ there is at least one occupied voxel. 
We define $F_{v} \subset F$ the set of non-occluded frontiers from the pose of the view candidate $v$, and $\delta_{v,f}$ the binary variable equals 1 if $f \in F_{v}$ and 0 otherwise. However, this definition of occlusion does not take into account the possible occlusion of unknown voxels along the ray that \textit{could} be occupied. 
We thus define the function $vis(v,f)$ estimating the visibility of the target frontier $f$ from the view candidate $v$, considering the potential occlusions due to unknown voxels, as follows: 
\begin{equation}
    vis(v,f) =  \delta_{v,f}\, e^{-u}
\end{equation}
with $u$ being the number of unknown voxels along the ray from the sensor pose $v$ to the target frontier $f$. Using these definitions, we define the information gain function $J_I(v)$ as:
\begin{equation}
    J_I(v) = \frac{1}{ \max\limits_{v \in V} |F_v|} \sum_{f \in \mathbf{F}} h_v(f) \, vis(v,f) 
\end{equation}
where 
\begin{equation} 
h_v(f) = \begin{cases}
1 & \text{if } |d(f,s_v)  - d^{*}| < \eta  \\
0.5 & \text{if }  d(f,s_v) < d^* - \eta  \\
0.5 \left(1-\frac{d(f,s_v)}{r_{max}}\right) & \text{if }   d(f,s_v) > d^* + \eta
\end{cases}
\end{equation}
The function $h$ prioritizes view candidates that observe frontiers from a distance close to $d^*$. Views at farther distances are penalized, as they do not satisfy the quality criteria. Closer viewing distances do satisfy the quality criteria, but have fewer voxels in the field of view.
Hence, they are also penalized, although less significantly.

\subsubsection{Navigation component}

During the exploration, a major fallback is to return to previously visited areas to observe a missed portion of the surface, generating additional and unnecessary travel. Furthermore, the position $s_v$ of the view candidate $v$ can be close in distance to the actual robot position $s_q \in \mathbb{R}^{3}$ but facing the opposite direction. Moving to that view would incur a costly change in direction with a trajectory that cannot be a straight line. To generate efficient trajectories, we include both of these aspects in the design of our navigation component $J_N$:
\begin{equation}
    J_N (q,v) = \psi(q,v) \frac{\min\limits_{v \in V} || s_v - s_q || }{|| s_v - s_q ||}
\end{equation}
where \begin{equation}
      \psi(q,v) = 1 - \frac{1}{\pi} \arccos{\left( \frac{ { vel }^T }{ || vel || } \cdot \frac{  s_v - s_q  }{  || s_v - s_q ||  }\right) }
\end{equation}
with $vel$ being the velocity vector of the robot. 

\subsection{Local Planning}

To represent the 3D space as a 3D map, we adopt ETHZ ASL's Voxblox representation \cite{voxblox}, which builds a TSDF and an ESDF concomitantly. The TSDF is used for the reconstruction reasoning and the ESDF by the navigation component to determine the freespace. We use the local planner from \cite{oleynikova2018safe} to generate dynamically feasible trajectories and replan in real-time to avoid obstacles. The local planner has two main components: first, the local trajectory optimizer that computes the best path between two waypoints without collisions. This trajectory is then smoothed out using the loco smoother \cite{oleynikova2018safe} to respect the velocity and acceleration constraints. Then, the intermediate goal selector creates sub-sections of the path between the current UAV pose and the goal and sends them one by one to the UAV's controller.

\begin{figure}[tb]
    \centering
    \includegraphics[width=\columnwidth]{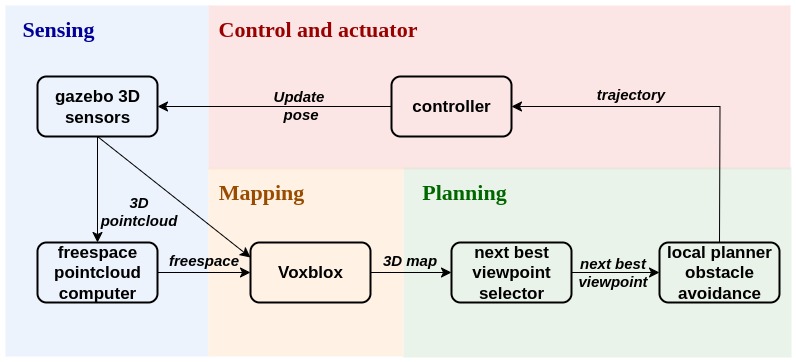}
    \caption{Overview of the main components of the proposed approach.}
    \label{fig:software_stack}
\end{figure}

To improve the UAV's navigation and obstacle avoidance, we have added a component that computes a freespace pointcloud from the UAV sensor's original pointcloud. First, the maximum range of the sensor is used to recreate a "full" pointcloud. Then, it is compared to the actual sensor input. Specifically, any point in the full pointcloud that lies in the same direction as a detected point from the sensor data but is farther away is removed. 
This freespace pointcloud gives additional information about the navigable environment perceived by the UAV and is passed to Voxblox to enhance the quality of the ESDF and TSDF. Making a difference between empty and unknown space is also crucial to assess occlusion within the information gain calculation. 

An overview of the global architecture, including the sensing, mapping, planning, and control components, is shown in Fig.~\ref{fig:software_stack}.

\begin{figure*}
    \centering
    \resizebox{0.835\textwidth}{!}{
    \begin{tikzpicture}
        \node[anchor=south west, inner sep=0] (image) at (0,0) {
            \includegraphics[width=2.0\columnwidth, valign=t]{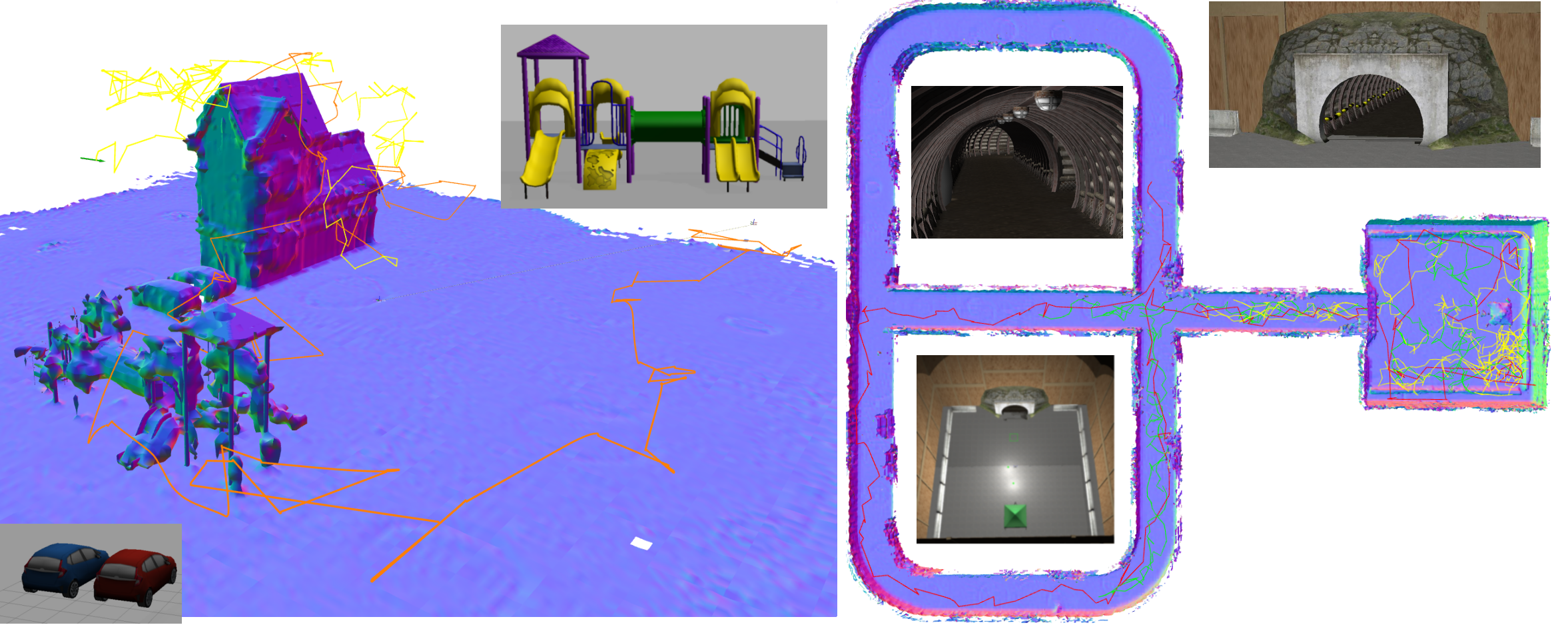}
        };

        \begin{scope}[x={(image.south east)}, y={(image.north west)}]

            \node[anchor=south east, fill=white, draw=black, rounded corners, font=\small] at (0.51, 0.04) {
                \begin{tabular}{@{}l@{}}
                    \textcolor{yellow}{\rule{10pt}{2pt}} Online-IPP \\
                    \textcolor{orange}{\rule{10pt}{2pt}} Ours \\
                \end{tabular}
            };

            \node[anchor=south east, fill=white, draw=black, rounded corners, font=\small] at (0.98, 0.02) {
                \begin{tabular}{@{}l@{}}
                    \textcolor{red}{\rule{10pt}{2pt}} Ours (average) \\
                    \textcolor{green}{\rule{10pt}{2pt}} Online-IPP (max) \\
                    \textcolor{yellow}{\rule{10pt}{2pt}} Online-IPP (min) \\
                \end{tabular}
            };

        \end{scope}
    \end{tikzpicture}
    }
    \caption{House and Tunnel environments (ground-truth mesh) with 15-minute trajectories.}
    \label{fig:trajectories}
\end{figure*}

\section{Numerical Experiments}
\label{sec:results}
In this section, we test and analyze our method in different simulated environments with varying characteristics.
To assess the efficiency of our approach, we compare its performance against three methods from the literature: an informative sampling method based on RTT* \cite{Schmid2020}, and two classical heuristic baselines.
The first baseline evaluates viewpoints by counting the number of visible frontiers from that position using raycasting. Intuitive but costly, it is adopted by several state-of-the-art approaches, as \cite{hardouin2023} and \cite{zhou2021}. The second one consists of always selecting the closest generated viewpoint as the next destination, which has consistently performed on-par with recent methods \cite{ufo_duberg}. 
For the view generation of these methods, we use the state-of-the-art method detailed in \cite{zhou2021}, which adopts a cylindrical sampling around the frontier.

\subsection{Simulation setup}

Performance evaluations were performed in ROS1-Gazebo simulation with a quadrotor UAV equipped with a 3D LIDAR providing a full 360-degree horizontal field of view (FOV) and a vertical FOV ranging from -35 to +30 degrees. The UAV model is based on an Intelaero UAV, using the same motor placement and dimensions. To guide the UAV in Gazebo, we designed a controller that follows the given trajectory perfectly. This perfect controller is used to prevent navigation bias when comparing the different approaches in simulation. 
We assume perfect localization as in all the major works of the field from the literature \cite{Schmid2020}, \cite{hardouin2023}, \cite{zhou2021}.

The experimental results presented in this paper took place in the following simulated worlds:
\begin{itemize}
    \item House: a 45x40x20m outdoor open field urban environment composed of a house, two cars, and a complex playground structure. All elements are dispersed within the map and disconnected, except for the ground surface. Object shapes are quite complex. 
    \item Tunnel: a 200x200x4m indoor structured terrain with a large cubic room and a narrow tunnel entry point. The tunnel splits into 3 sections that remerge further. This world requires a more precise navigation and combines an almost open area and narrow corridors. 
\end{itemize} 
These two environments have been selected for their challenging and deeply different characteristics. 
The scattered objects of interest in "House" and the combination of both a wide open room and multiple branching narrow corridors in "Tunnel" create complex topologies where the robot must efficiently prioritize the views to select. Typically, in the "House" environment, the empty ground surface should be covered quickly to focus on the more complex geometry of the objects within the scene. In "Tunnel", the variety within the environment requires a strategy that can efficiently handle both wide and tight spaces for view selection and navigation.   
Additionally, the spaced objects in "House" and the scale of "Tunnel" can easily induce unnecessary back-and-forth motions. 
The mission time has been limited to 15 minutes for both environments.
This time limit is intentionally short in order to evaluate the ability of the approaches to prioritize regions of interest within a limited time frame. 

Across all experiments and scenes, we keep the same values of $\alpha = \beta = 0.5$. Other simulation parameters are detailed in Table \ref{tab:simulation_parameters}. 

\begin{table}[ht]
\centering
\begin{tabular}{|l|c|}
\hline
\textbf{Parameter} & \textbf{Value}   \\
\hline
Robot Radius           & 1.0 m   \\
Voxel Size $d_s$             & 0.25 m \\
Lidar Min/Max Range $r_{min}, r_{max}$        & 0.42m / 10m   \\
Vertical Min/Max Angle     & -35° / 30°  \\
Maximum velocity $v_{max}$             & 4.5 m/s       \\ 
Maximum absolute acceleration $a_{max}$     & 4.8 m/s²   \\ 
\hline
\end{tabular}
\caption{Simulation parameters.
\vspace{-.6cm}}
\label{tab:simulation_parameters}
\end{table}

\subsection{Metrics}

The quality of a reconstructed map is judged on its completeness and its fidelity.  
For this reason, we divide our evaluation into two steps. First, we compare our solution to the other methods with respect to coverage and speed. Namely, we rely on the following three metrics: coverage ratio, distance traveled, and mean map error. 
To evaluate coverage, we compare the TSDF map obtained at the end of the execution with a TSDF ground truth containing all the observable surfaces, obtained through a manual traversal of the environments. We extract all the surface voxels of the ground-truth map in the set  $ \mathcal{M}_{surface}$. A voxel in $ \mathcal{M}_{surface}$ is covered if the distance value difference between the same voxel in the ground-truth and the evaluated map is less than the voxel diagonal. The map error is the average distance error across all surface voxels of the ground truth map.  

The second part of our analysis focuses on map accuracy and is based on two metrics: the function $Z(M)$ expressed in eq.~(\ref{quality_function}), and the percentage distance $\Delta$ between the value of the objective function $Z_{sat}(M;w^*)$ corresponding to the final map, and its optimal value $Z^*$. 
In both cases, the completeness factor is approximated to $ \xi (M) = |M_S| / \mathcal{M}_{surface}$.
Since none of the methods is deterministic, each experiment has been repeated five times, and the averaged values are reported in Tables \ref{tab:coverage_comparaison}, \ref{tab:house_quality}, and \ref{tab:tunnel_quality}.

\begingroup
\setlength{\tabcolsep}{5pt}
\begin{table*}[h]
\centering
\begin{tabular}{|l|c|c|c|c|c|c|c|c|}
  \hline
  \makecell{\textbf{Methods}\\ } & \makecell{\textbf{House}\\ \textbf{Coverage}} &
  \makecell{\textbf{Tunnel}\\ \textbf{Coverage}} &
 \makecell{\textbf{House}\\ \textbf{Path length}} &
  \makecell{\textbf{Tunnel}\\ \textbf{Path length}} &
    \makecell{\textbf{House}\\ \textbf{Map Error}} &
  \makecell{\textbf{Tunnel}\\ \textbf{Map Error}} &
    \makecell{\textbf{House}\\ \textbf{$Z$}} &
  \makecell{\textbf{Tunnel}\\ \textbf{$Z$ }} \\
  \hline
    Ours ($d^*=4.5$m)         & 9507 (86.3\%) & 54850 (71.5\%) & 228.6 & 411.8 & 3,74\% & 4,84\% & 0.83 & 0.71 \\
    Online-IPP \cite{Schmid2020}            & 8681 (78.8\%) & 22865 (29.8\%) & 731 & 819  & 3,64\% & 5,11\%  &0.75 &0.3\\
    Closest frontier (small spread)            & 7881 (71.6\%) & 29437 (38.4\%)  & 195.6 & 233.8 & 3,85\% & 5,16\%  &0.69 &0.38 \\
    Closest frontier (large spread)            & 8221 (74.7\%) & 18950 (24.71\%)  & 242.4 & 163 & 3.66\% & 4.01\%  &0.72 &0.24 \\
    Frontier count (small spread)     & 7857 (71.4\%) & 26822 (34.9\%)  & 230.2 & 187.8 & 3,63\% & 4,66\%  &0.68 &0.35 \\
        Frontier count (large spread)     & 9377 (85.21\%) & 26266 (34.25\%)  & 245.8 & 182.66 & 3,75\% & 4,78\%  &0.83 &0.34  \\
  \hline
\end{tabular}
\caption{Performance comparison across different environments.
\vspace{-0cm}}
\label{tab:coverage_comparaison}
\end{table*}
\endgroup

\begin{table*}[h]
\centering
\begin{tabular}{|l|c|c|c|c|c|c|}
  \hline
  \makecell{ \textbf{Methods (House) }} & 
  \makecell{$\Delta(\sqrt{1 / Z^*}=1)$ } &
    \makecell{$\Delta(2)$ } &
  \makecell{$\Delta(4)$} &
    \makecell{$\Delta(5)$} &
  \makecell{$\Delta(8)$} &
    \makecell{ \textbf{Coverage}} \\
  \hline
    Ours ($d_{obs}=1.5\pm 0.5$m)         & 
     1.33\% & 0.19\%  & $\leq 0.01\%$ & $\leq 0.01\%$  & $\leq 0.01\%$ & 69.37\% \\
    Ours ($d_{obs}=4.5\pm 0.5$m)   & 
    3.79\% & 0.79\% & 0.01\% & $\leq0.01\%$ & $\leq 0.01\%$ & 86.39\% \\
    Ours ($d_{obs}=8.5\pm 0.5$m)           & 
    13.93\% & 3.76\% & 0.33\% & 0.07\% & $\leq 0.01\%$ & 99.62\%  \\
    Online-IPP \cite{Schmid2020}        & 4.77\% & 1.07\% & 0.07\% & 0.01\% &  $\leq 0.01\%$ & 78.8\%  \\
    Frontier count (large spread)      & 
    2.75\% & 0.52\% & 0.01\% & $\leq0.01\%$ & $\leq 0.01\%$ & 74.7\%  \\
    Frontier count (small spread)      & 
    5.14\% & 1.18\% & 0.04\% & $\leq0.01\%$ & $\leq 0.01\%$ & 71.4\%  \\
    Closest frontier (large spread)      & 
    4.03\% & 0.67\% & 0.01\% & $\leq0.01\%$ & $\leq 0.01\%$ & 85.21\%  \\
        Closest frontier (small spread)      & 
    3.99\% & 0.75\% & $\leq0.01\%$ & $\leq0.01\%$ & $\leq 0.01\%$ & 71.6\%  \\
  \hline
\end{tabular}
\caption{Distance to quality objectives for each approach in the house environment.
\vspace{-0cm}}
\label{tab:house_quality}
\end{table*}

\begin{table*}[h]
\centering
\begin{tabular}{|l|c|c|c|c|c|c|}
  \hline
  \makecell{ \textbf{Methods (Tunnel) }} & 
  \makecell{$\Delta(\sqrt{1 / Z^*}=1)$ } &
    \makecell{$\Delta(2)$ } &
  \makecell{$\Delta(4)$} &
    \makecell{$\Delta(5)$} &
  \makecell{$\Delta(8)$} &
    \makecell{ \textbf{Coverage}} \\
  \hline
    Ours ($d_{obs}=1.5 \pm 0.5$m)          & 
    0.85\% & 0.13\%  & $\leq0.01\%$ & $\leq0.01\%$ & $\leq 0.01\%$ & 41.58\%  \\
    Ours ($d_{obs}=4.5 \pm 0.5$m)    & 
    0.26\% & 0.02\% & 0.01\% & 0.01\% & $\leq 0.01\%$ & 71.5\%  \\
    Online-IPP  \cite{Schmid2020}  &     0.56\% & 0.04\% & $\leq0.01\%$ & $\leq0.01\%$ & $\leq 0.01\%$ & 29.8\%  \\
    Frontier count (large spread)      & 
    0.71\% & 0.06\% & $\leq0.01\%$ & $\leq0.01\%$ & $\leq 0.01\%$ & 34.25\%  \\
    Closest frontier (large spread)      & 
    1.15\% & 0.14\% & $\leq0.01\%$ & $\leq0.01\%$ & $\leq 0.01\%$ & 24.71\%  \\
  \hline
\end{tabular}
\caption{Distance to quality objective for each approach in the tunnel environment.
\vspace{-0cm}}
\label{tab:tunnel_quality}
\end{table*}

\subsection{Results}

\subsubsection{Performance comparison}
The first conclusion that can be drawn from the first set of comparative results presented in Table~\ref{tab:coverage_comparaison} is that our proposed approach outperforms the other three methods in terms of coverage for both environments while maintaining a similar mean error. In "Tunnel" notably, our approach covers 86\% more surface voxels than the second-best approach. The performance differences are less pronounced in "House", which is closer to a standard open environment, but rise drastically in "Tunnel". This is due to the fact that, in our simulations, all the other methods struggle transitioning from the wide room to the narrow tunnel shaft and lose a large amount of time (see Fig. \ref{fig:trajectories}). Random sampling methods do not manage to efficiently cover some hard-to-see surfaces, located at the bottom of the room walls, highlighting the benefits of surface-based view generation. It should be noted that for Online-IPP \cite{Schmid2020}, there is significant variability between instances. In most of our experiments, it hardly enters the tunnel shaft and remains in the main room for most of the mission time (see Fig. \ref{fig:trajectories}). However, when it manages to find the tunnel entrance, its speed of exploration in the narrow corridor is unmatched by the other methods, including ours, despite significant back and forth. 
Our approach is the most consistent, as our high coverage ratio demonstrates, and with the best path efficiency overall in all environments (see Fig. \ref{fig:trajectories}, Table \ref{tab:coverage_comparaison}) across all methods. In particular, our paths are 3 to 4 times shorter than Online-IPP's and provide better coverage. 
We can finally note that a quality objective corresponding to half the field of view of the sensor proves to be a good balance between speed and accuracy, as demonstrated by the values of $Z$ in Table \ref{tab:coverage_comparaison}.

\subsubsection{Adaptation to the quality objective}

For the second part of our analysis, we consider the percentage distance $\Delta$ to different quality objectives obtained using different observation distance intervals $d_{obs}\pm \eta$: close-range (1.5m), middle-range (4.5m), and high-range (8.5m) with $\eta$ always 0.5m.
We also measure the performance of two versions of closest frontiers and frontiers count: one with a large sampling range of 3 to 8 meters for view generation, and a version with a smaller range of 4 to 5 meters.  
These results are presented in Tables \ref{tab:house_quality} and \ref{tab:tunnel_quality} for the House and Tunnel environment respectively. For readability purposes, and given its direct relation with $d^*$, we use the value of $\sqrt{1/Z^*}$ instead of $Z^*$ to specify the considered quality objective.

When the observation distance increases, our solution coverage increases significantly: in "House", the best-performing method in terms of coverage is ours, with a high range. In "Tunnel", ours-middle-range is also the top performer (we did not evaluate a high-range as the narrow corridors of the tunnel do not allow it).
However, as expected, when observing from higher ranges, the accuracy of the observed map worsens. Conversely, with the smallest range of observation, our exploration is slower, but the map fidelity increases. 
In "House", ours ($d^* \in [1;2]$), is slower but has the best accuracy of all the strategies with an error 2 to 4 times inferior to the other methods (ours excluded). For higher observation ranges, our method always has a distance to the quality objective of 0.01\% or less, but with the corresponding confidence value being very small, the comparison between strategies appears less significant. In "Tunnel", ours-close-range ($d_{obs} \in [1;2]$) has a higher distance to the quality target than ours-middle-range ($d_{obs} \in [4;5]$). The small value of $d_{obs}$ in the narrow corridors causes big angle changes between views, which generates a higher error. In general, we can notice that our approach is either the fastest to cover the environments or the one that achieves the best accuracy, depending on the corresponding quality target.

\section{Conclusion}
In this work, we presented a novel, quality-guided framework for exploring and reconstructing unknown environments using an aerial robot equipped with a range sensor. The proposed NBV approach enables the robot to adapt its selection of future observation points based on the perceived environment and user-specified reconstruction fidelity requirements, achieving a more efficient and task-driven exploration. We tested the proposed approach in a series of experiments involving diverse environments and quality objectives and compared it to state-of-the-art standard alternatives. Our extensive results showed that our method effectively adjusts exploration efficiency and reconstruction accuracy while consistently outperforming alternative methods in both map coverage and final accuracy, demonstrating the value of incorporating quality criteria into the design of planning algorithms.

In the future, we plan to expand our approach to guide multiple UAVs as a team to obtain scalable and cooperative solutions for exploring and reconstructing large-scale, complex environments.

\bibliographystyle{IEEEtran}
\bibliography{biblio}

\end{document}